\documentclass{article}

\usepackage{arxiv}

\usepackage[utf8]{inputenc} % allow utf-8 input
\usepackage[T1]{fontenc}    % use 8-bit T1 fonts
\usepackage{hyperref}       % hyperlinks
\usepackage{url}            % simple URL typesetting
\usepackage{booktabs}       % professional-quality tables
\usepackage{amsfonts}       % blackboard math symbols
\usepackage{nicefrac}       % compact symbols for 1/2, etc.
\usepackage{microtype}      % microtypography
\usepackage{lipsum}		% Can be removed after putting your text content
\usepackage{graphicx}
\usepackage{natbib}
\usepackage{doi}
\usepackage{amsmath}

\title{Look Before You Leap! Designing a Human-Centered AI System for Change Risk Assessment}

\date{} 				% Or removing it

\author{Binay Gupta\\
	Walmart Global Tech\\
	Bangalore, India\\
	\texttt{binay.gupta@walmart.com}\\
	\And
	Anirban Chatterjee\\
	Walmart Global Tech\\
	Bangalore, India\\
	\texttt{anirban.chatterjee@walmart.com}\\
	\And
	Harika Matha\\
	Walmart Global Tech\\
	Bangalore, India\\
	\texttt{matha.harika@walmart.com}\\
	\AND
	\href{https://orcid.org/0000-0002-0605-630X}{\includegraphics[scale=0.06]{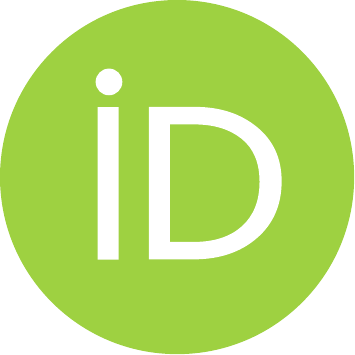}\hspace{1mm}Kunal Banerjee}\\
	Walmart Global Tech\\
	Bangalore, India\\
	\texttt{kunal.banerjee1@walmart.com}\\
    \And
    Lalitdutt Parsai\\
    Walmart Global Tech\\
	Bangalore, India\\
	\texttt{lalitdutt.parsai@walmart.com}\\
	\And
	Vijay Agneeswaran\\
	Walmart Global Tech\\
	Bangalore, India\\
	\texttt{vijay.agneeswaran@walmart.com}\\
}

\renewcommand{\shorttitle}{Designing a Human-Centered AI System for Change Risk Assessment}

\hypersetup{
pdftitle={A template for the arxiv style},
pdfsubject={q-bio.NC, q-bio.QM},
pdfauthor={David S.~Hippocampus, Elias D.~Striatum},
pdfkeywords={First keyword, Second keyword, More},
}

\begin{document}
\maketitle

\begin{abstract}
Reducing the number of failures in a production system is one of the most challenging problems in technology driven industries, such as, the online retail industry.
To address this challenge, change management has emerged as a promising sub-field in operations that manages and reviews the changes to be deployed in production in a systematic manner. However, it is practically impossible to manually review a large number of changes on a daily basis and assess the risk associated with them.
This warrants the development of an automated system to assess the risk associated with a large number of changes. There are a few commercial solutions available to address this problem but those solutions lack the ability to incorporate domain knowledge and continuous feedback from domain experts into the risk assessment process.
As part of this work, we aim to bridge the gap between model-driven risk assessment of change requests and the assessment of domain experts by building a continuous feedback loop into the risk assessment process. Here we present our work to build an end-to-end machine learning system along with the discussion of some of practical challenges we faced related to extreme skewness in class distribution, concept drift, estimation of the uncertainty associated with the model's prediction and the overall scalability of the system.
\end{abstract}

% keywords can be removed

\section{Introduction}
In any technology driven industry, launch of a new business or launch of new product features for an existing business to customers requires a series of software changes to a base system that is already in production. Each of these changes carries with it some likelihood of failure. Reducing the number of failures in a production system is one of the key challenges. It is especially important in the current era of agile development that has a tight delivery schedule.
%The situation may be further exacerbated when there is a large volume of changes, which are generally considered to be low in risk, severely restrict thorough inspection and review before deployment.
The situation may be further exacerbated when there is a large volume of changes, which severely restricts thorough inspection and review before deployment.
From our experience, another impediment in manual change risk assessment occurs when the risk for a change is marked as ``low'' by the change requester 
%(which need not be so, especially, if the developer is new or less skilled and hence had applied poor judgement) 
-- such requests are often ignored altogether by domain experts for reviews, and these may manifest as critical issues later in the pipeline. 
In fact, in the context of Walmart, we observe that a substantial percentage\footnote[1]{\label{fn}{We abstain from providing the exact numbers to maintain confidentiality.}} of major production issues occur due to planned and so-called ``low-risk'' changes in e-commerce market and US stores.
The monetary impact of these major issues is also quite significant\textsuperscript{\ref{fn}}.
The number of such changes per week is so high\textsuperscript{\ref{fn}} on average that it is practically impossible to manually review all these change requests due to limited bandwidth of the human experts.
This necessitated the development of an automated risk assessment system for change requests. 
\par In this paper, we present our experience of exploring the following questions  while building an automated risk assessment system:
\begin{itemize}
    \item Can we reliably build a failure probability model for changes which can provide actionable insights from the change data to the change management team?
    \item Can we optimally seek feedback from the domain experts for the model's inference on a limited number of changes so that it improves the model's performance as well as does not over-burden the domain experts with feedback requests?
\end{itemize}
The remainder of this paper is organized as follows.
In Section~\ref{sec:problem}, we provide an overview of the problem.
In Section~\ref{sec:data}, we provide a brief description of our dataset.
In Sections~\ref{sec:system} \& \ref{sec:deployment}, we elaborate on the end-to-end system and its deployment, respectively.
In Sections~\ref{sec:performance} \& \ref{sec:observations}, we talk about the business impact of this solution and some of the interesting observations we made in the course of this work, respectively, and finally, we conclude this paper in Section~\ref{sec:conclusion}.

\section{Problem Overview}\label{sec:problem}
Our main goal is to determine if we can predict the probability that a change will cause a major production issue based on the information available for that change request.
Prediction at an earlier stage is likely to be much less precise, and prediction at a later stage would be much less useful because fewer options would be available to mediate the risk.

% \subsection{Challenges}

% One of the major problem we have faced is imbalance class distribution in the data. It makes machine learning model bias towards predicting majority class, which in turn leads to high false negative rate and monetary loss. Also, during holiday period, number of change request drops sharply and associates avoid pushing risky changes in production. This creates cyclical shift in data distribution. In addition to that, process change in operation, formation or merger of teams lead to gradual concept drift. Along with these difficulties, running the system in production seamlessly on large amount of data makes the problem even more challenging. 
% \subsection{Contribution}
% \begin{itemize}
% \item Identifying change requests which may got converted to major incident not only reduces repetitive task of change requester board, but also helps them to focus on non-repetitive and creative aspect of the problem like change quality issue.
% \item Also, our model presents uncertainty score against each prediction which makes the job of human expert even more easier. 
% \item We consider domain experts' feedback and present explanation for model's decision to gain trust of our client.
% \end{itemize}

\section{Data Description}\label{sec:data}
We have collected change request data for one year.
Each instance in this data consists of several attributes or features.
We can logically divide them into four primary categories:
\begin{itemize}
\item {\textbf{Descriptive Feature}}: These are plain text information about the change, such as, change summary, change description, and a few others.
\item {\textbf{Q \& A Feature}}: These are the answers provided by the change requester to a set of predefined questions, such as, 
%``what is the risk level for this change'', ``what is risk of this change'',
``whether the change was previously implemented or not'', etc.
\item {\textbf{Team Profile}}: This information is not readily available with the change data but we derive it from the historical data. These features primarily reflect the tendency of a team to raise change requests which create major issues in production.
\item {\textbf{Change Importance}}: These features reflect the perception of the change requester regarding the impact, importance and the risk of a change request. 
\end{itemize}

We also associate labels with every change data instance.
We associate the changes, which did not create any major production issue, with the label ``normal''.
We label the others as ``risky''.
Our training sample consists of 600K change instances among which only 540 belong to the class ``risky'', which is only $0.09\%$ of all the change instances.

\section{Risk Assessment System}\label{sec:system}
\begin{figure*}[htb]
    \centering
    \includegraphics[scale= 0.4]{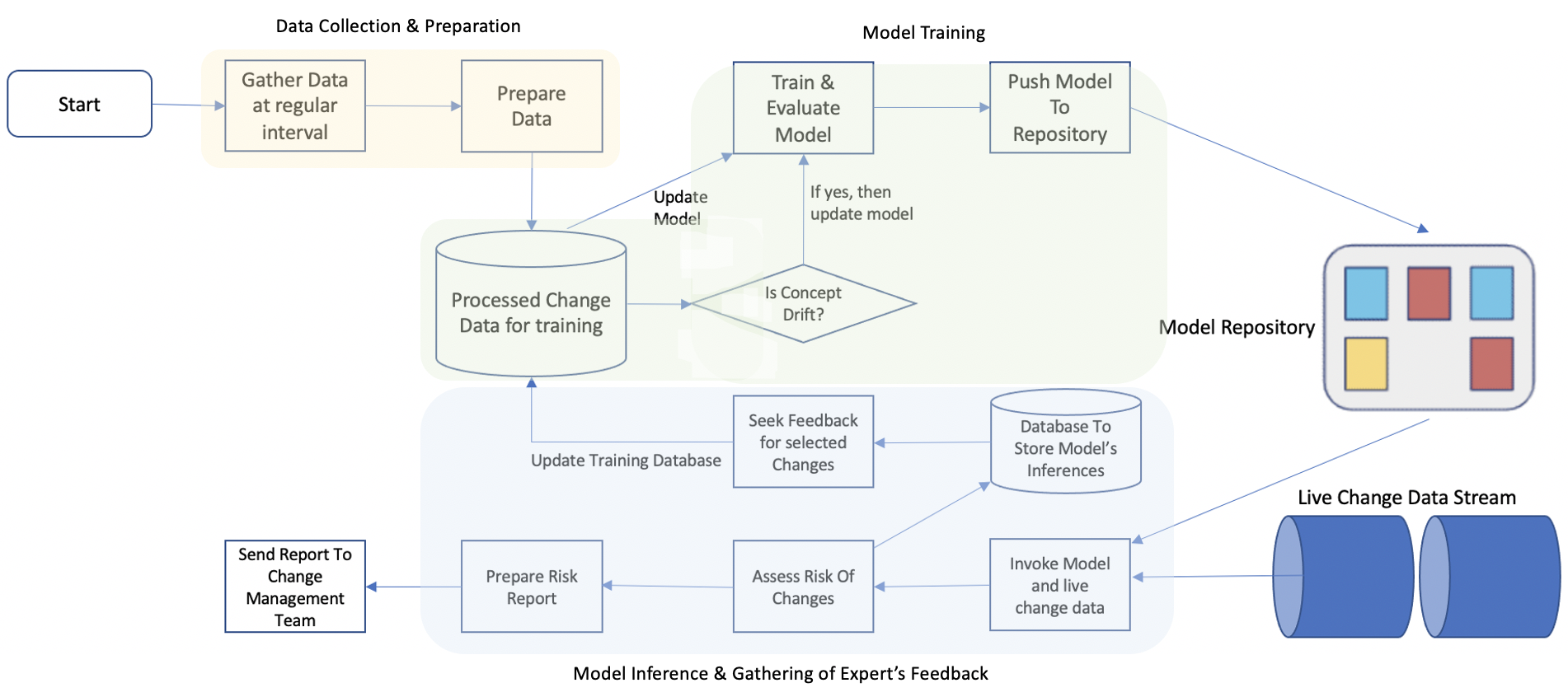}
    \caption{Conceptual diagram of end-to-end system workflow.}
    \label{fig:conD}
\end{figure*}

We build an automated risk assessment system which has conceptually three main functional components:
\begin{itemize}
    \item data collection and preparation,
    \item model training and monitoring,
    \item model inferencing and gathering of expert's feedback.
\end{itemize}
Figure~\ref{fig:conD} illustrates a conceptual diagram of our end-to-end system workflow.
We explain each and every functional component of the system in the following subsections.

\subsection{Data Collection and Preparation}
In this part of the system, we collect change related data from multiple sources and aggregate them.
Once aggregated, we prepare the training data for the subsequent training stage.
It is important to mention here that we pose this task as a classification problem with a high degree of class imbalance.
A subset of features that we use for training the classification model are raw attributes of the change requests, and such change attributes are readily available in change data that we collect.
However, some of the features that we feed to the machine learning (ML) model are derived features, such as, the features to indicate the degree of severity expressed in the change description or a team's tendency to cause major production issues through changes, and many others.
As the data exhibit high degree of class imbalance, we resort to up-sampling the minority class by synthesizing data instances.

% Next, during processing phase we impute missing values, encode categorical features, upsample instances from minority class to generate final training data. Linear regression is applied to impute missing values. Both label encoding and OHE is used for categorical features judiciously. Brute force up-sampling of minority class can cause over-fitting. Similarly, we may end up discarding potentially useful information if we randomly downsample instances from majority class. To mitigate both the problem, we have used a modified version of smote (M-SMOTE) where only safe samples from minority class are used to synthesis new instances, thus we can avoid generating noisy and boundary samples.

\subsection{Model Training and Monitoring}
We use a gradient-boosted decision tree (XGBoost)~\cite{Chen} to generate the probability with which a new change request may cause major issues in production and this ML model is at the core of this system. 
We consider this probability as the estimation of risk for a change.

\subsubsection{\textbf{Concept Drift}}
We generally train the model once in a month.
However, we have a system in place to monitor any significant shift in data pattern which may substantially degrade the performance of the model (see Figure~\ref{fig:conD}).
In case we detect any such drift, we initiate an out-of-cycle training of the model with the latest change data.
This kind of drift in data pattern is called \textit{concept drift} and is formally defined as follows:
\begin{equation}
%\begin{aligned}
\exists X: p_{t0}(X, y) \neq p_{t1}(X, y)
%\end{aligned}
\end{equation}
This definition explains concept drift as the change in the joint probability
distribution for input $X$ and prediction $y$ between two time points $t0$ and $t1$.

% During model training phase, we have tried both supervised and unsupervised models before deciding on which algorithm will perform best. One class SVM and isolation forest are the two algorithm we have used from unsupervised classification paradigm. On supervised learning algorithm, we have analysed performance of logistic regression, XGBoost and Deep Neural Network. \par
% Unsupervised machine learning algorithms are useful in absence of class label and it always under-perform supervised learning method. Logistic Regression assumes linear relationship between dependent and independent variable which may not always hold true. Gradient boosting and Deep Neural Network emerge as clear winner as they can learn complex function better. As XGBoost is less resource intensive and explaining model's decision is easier here, we have decided to go with XGBoost. While choosing best set of hyper-parameter, we have used Bayesian HPO technique.

\subsubsection{\textbf{Detection Of Concept Drift}}
We use a modified form of \textit{Kolmogorov-Smirnov (KS) Test} to detect concept drift in data.
Before we introduce how we apply it in this context, we first briefly review the standard form of KS Test.
\par Suppose we have two samples $A$ and $B$ containing univariate observations.
We would like to know with a significance level of $\alpha$, whether we can reject the null hypothesis that the observations in $A$ and $B$ originate from the same probability distribution.
If no information is available regarding the data distribution, it is safe to assume that the drawn observations are i.i.d., we can use the rank-based KS test to verify the proposed hypothesis.
According to it, we can reject the null hypothesis at level $\alpha$ if the following inequality is satisfied:
\begin{equation}
%\begin{aligned}
D > c(\alpha)\sqrt{\frac{n + m}{nm}}
%\end{aligned}
\end{equation}
where the value of $c(\alpha)$ can be retrieved from a known table, $n$ is the number of observations in $A$ and $m$ is the number of observations in $B$. The right side of the inequality is the target $p$-value. $D$ is the Kolmogorov-Smirnov statistic, i.e.,
the obtained $p$-value, and is defined as follows:
\begin{equation}
%\begin{aligned}
D = \sup_x | F_{A}(x) - F_{B}(x) |
%\end{aligned}
\end{equation}
where
\begin{equation}
%\begin{aligned}
F_{A}(x) = \frac{1}{| A |}\sum_{a \in A, a \leq x} 1, \  F_{B}(x) = \frac{1}{| B |}\sum_{b \in B, b \leq x} 1
%\end{aligned}
\end{equation}
$F(\cdot)$ represents cumulative distribution function. We note that $D$ can actually be computed as follows:
\begin{equation}
%\begin{aligned}
D = \max_{x \in A \cup B} | F_{A}(x) - F_{B}(x) |
%\end{aligned}
\end{equation}
\par In order to quantify drift we use a modified version of KS algorithm. We first measure the drift in each and every feature and later we combine them using weighted average. More formally, we compute the final drift between two multi-variate samples of data as follows:
\begin{equation}
%\begin{aligned}
D_{final} = \frac{1}{K}\sum_{i=1}^{K} w_i D_i
%\end{aligned}
\end{equation}
where $D_i$ is the measured drift in $i^{th}$ feature between the two samples according to KS algorithm and $w_i$ is the importance of $i^{th}$ feature as computed by XGBoost while training and $K$ is the total number of features.
\par Once the value of $D_{final}$ crosses a certain threshold, it raises an alarm to update the model by retraining. While training the model, we assign higher weights to more recent data points so that the model is more tuned to the latest pattern in the dataset.

\subsection{Model Inferencing and Gathering of Expert's Feedback}
This part of the system is responsible for ingesting the live change data in batches into the system, running the latest version of the model against them to generate the risk score and sending back the risk report to the change management team.
\par It is also responsible for gathering expert's feedback on a small sample of changes. It seeks an expert's feedback only for those changes for which the model exhibited a high degree of uncertainty. It actually ranks all the change requests in a batch according to their estimated uncertainty of prediction and sends top $m$ change requests to experts for feedback. The subsection below provides a brief overview of how we estimate the predictive uncertainty of the model.

%\subsubsection{\textbf{Estimation Of Predictive Uncertainty}}
\noindent \textit{\textbf{Estimation Of Predictive Uncertainty.}}
While predictive uncertainty is widely studied for deep learning based models~\cite{dluncertainty},~\cite{Gal}, the topic seems to be under-explored for gradient boosting based models such as XGBoost. We estimate the uncertainty associated with the predictions of the model within standard Bayesian ensemble based framework~\cite{Gal}. Total uncertainty is caused by both \textit{data uncertainty} and \textit{knowledge uncertainty}. Conceptually, we express the uncertainty associated with a prediction of the model as the mutual information between model parameters $\boldsymbol{\theta}$ and prediction $y$. We can estimate the mutual information between model parameters $\boldsymbol{\theta}$ and prediction $y$ as given below~\cite{Malinin}:
\begin{equation}
\begin{aligned}
\mathcal{I}[y, \boldsymbol{\theta} \rvert \mathbf{x}, \mathcal{D}] &= \mathcal{H}[P(y \rvert \mathbf{x}, \mathcal{D})] - \mathbb{E}_{p(\boldsymbol{\theta} \rvert \mathcal{D})}[\mathcal{H}[P(y \rvert \mathbf{x}, \boldsymbol{\theta})]] \\
&\approx \mathcal{H}[\frac{1}{M} \sum_{m=1}^M P(y \rvert \mathbf{x}; \boldsymbol{\theta}^{(m)})] - \frac{1}{M}\sum_{m=1}^M \mathcal{H}[P(y \rvert \mathbf{x}; \boldsymbol{\theta}^{(m)})]
\end{aligned}
\end{equation}

Here, $\mathbf{x}$ represents the feature-set corresponding to the prediction $y$,  $\mathcal{D} = \{\mathbf{x}^{(i)}, y^{(i)}\}_{i=1}^N$ represents the entire dataset and $M$ is the total number of trees constructed by XGBoost. This is expressed as the difference between the entropy ($\mathcal{H}$) of the predictive posterior, a measure of \textit{total uncertainty}, and the expected entropy of each model in the ensemble, a measure of expected \textit{data uncertainty}. Their difference is a measure of ensemble diversity and estimates knowledge uncertainty.
\begin{figure*}[ht!]
  \centering
  \begin{minipage}[b]{0.45\textwidth}
    \hspace{1 cm}
    \includegraphics[scale=0.45]{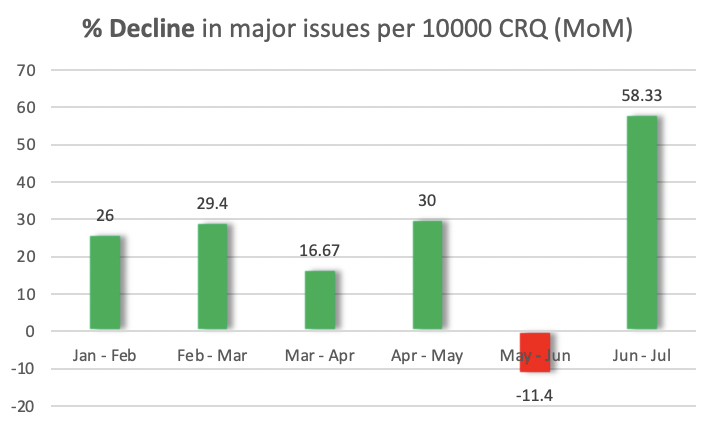}
    \caption{Percentage decline in major production issues (CRQ = change requests, MoM = month-over-month).}
    \label{fig:MI}
  \end{minipage}
  \hfill
  \begin{minipage}[b]{0.45\textwidth}
    \hspace{1cm}
    \includegraphics[scale=0.45]{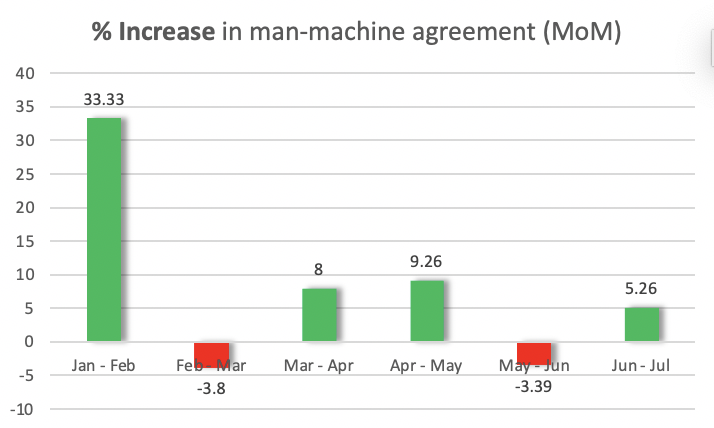}
    \caption{Percentage improvement in man-machine agreement (CRQ = change requests, MoM = month-over-month).}
    \label{fig:agreement}
  \end{minipage}
\end{figure*}
\section{Deployment and Monitoring}\label{sec:deployment}
We deploy the entire system as a workflow on an internal machine learning platform. Currently, it processes around 60$K$ change request per week. We have a dashboard in place to monitor several metrics related to the business impact of the system. The dashboard gets updated as soon as new data comes in. We build the pipeline for drift detection and the subsequent retraining of the model, as required, using MLFlow~\cite{mlflow}.

\section{Model Performance and Business Impact}\label{sec:performance}
\subsection{Model's performance}
We explore multiple options such as one-class SVM, isolation forest, logistic regression, deep neural network and XGBoost, to identify change requests with high risk. We consider true positive rate (TPR) and  false positive rate (FPR) as the performance metrics for the models. As Table~\ref{tab:comparison} suggests, deep neural network and XGBoost exhibit much better performance than the other methods we explored. To choose between XGBoost and deep neural network, we compute the positive likelihood ratio and XGBoost emerges the winner with respect to this metric. We computed all these metrics to evaluate a model's performance against validation dataset.

\subsection{Business Impact}
We primarily monitor two metrics to keep track of the business impact: \textit{number of major issues per 10000 CRQ (change requests)} and \textit{percentage of man-machine agreement}.
\par Figure~\ref{fig:MI} represents the month-over-month (MoM) improvement in the number of major issues per 10000 CRQ from January, 2021 to July, 2021. We observe around 85\% decline in this metric in July, 2021 with respect to January, 2021\footnote[2]{\label{fn2}We provide the relative variations of this metric MoM. Absolute values of this metric are confidential.}. We attribute the slight increase in this metric in June with respect to May to concept drift in data but we could reverse this trend by proactive detection of concept drift and subsequent retraining of the model.
\par \textit{Percentage of man-machine agreement} is a metric which represents the percentage of high risk changes as predicted by the model, which have actually been accepted as the high risk changes by domain experts. It is primarily an indicative of the confidence of business on this predictive system. Figure~\ref{fig:agreement} represents month-over-month improvement in this metric from January, 2021 to July, 2021\textsuperscript{\ref{fn2}}. Observe a slight dip in this metric in March and June with respect to February and May respectively. However, this trend has never lasted because of the continuous gathering of feedback from domain experts and incorporating the same into the model.
% We attribute this upward trend to continuous gathering of feedback from domain experts and incorporating the same into the model along with incorporating domain knowledge into the model as features.

%Recently, the production team has confirmed that the net savings realized by preventing major incidents caused due to changes ranges into multi-million dollars as in Q2 of 2021.
%Note that there may be other factors (e.g., software design changes) that have contributed to the savings; however, it is acknowledged that our AI based prediction system has definitely played a key role.

% \begin{center}
%   \begin{tabular}{ | l | c | c| }
%     \hline
%     \textbf{Algorithm} & \textbf{True Positive Rate} & \textbf{False Positive Rate} \\ \hline
%     One Class SVM & 52.7 & 18.6 \\ \hline
%     Isolation Forest & 51.3 & 18.9 \\ \hline
%     Logistic Regression & 62.5 & 14.5 \\ \hline
%     XGBoost & 78.9 & \textbf{9.3} \\ \hline
%     Deep Neural Network & \textbf{79.1} & 9.7 \\ \hline
%   \end{tabular}
% \end{center}
\begin{table}
\centering
%\captionsetup{labelformat=empty}
\caption{Comparative Analysis of ML Algorithms}
\label{tab:comparison}
\begin{tabular}{lrrr} 
\toprule
\textbf{Algorithm}       & \textbf{TPR(\%)} & \textbf{FPR(\%)} & \textbf{Positive Likelihood }\\
& & & \textbf{Ratio (TPR/FPR)}  \\ 
\midrule
One Class SVM            & $52.7 \pm 0.01$          & $18.6 \pm 0.01$        & $2.83$ \\ 
%\hline
Isolation Forest         & $51.3 \pm 0.03$          & $18.9 \pm 0.03$        & $2.71$ \\ 
%\hline
Logistic Regression (LR) & $62.5 \pm 0.01$          & $14.5 \pm 0.01$        & $4.31$ \\ 
%\hline
Deep Neural Network      & $\mathbf{79.1 \pm 0.02}$ & $9.7 \pm 0.02$         & $8.15$ \\ 
%\hline
XGBoost                  & $78.9 \pm 0.01$          & $\mathbf{9.1 \pm 0.01}$ & $\mathbf{8.67}$ \\
\bottomrule
\end{tabular}
\end{table}
\section{Some Observations}\label{sec:observations}
We share some of the interesting observations we made while building this system and how we dealt with them.
\subsection{\textbf{Up-sampling Minority Class}} We observe a significant variablity (see Table~\ref{tab:upsampling}) in model's performance with different up-sampling methods.
\begin{table}[ht!]
\centering
\caption{Experiments With Different Up-sampling Techniques in Learning By Oversampling}
\label{tab:upsampling}
\begin{tabular}{lrr} 
\toprule
\textbf{XGBoost with Different Up-sampling Methods} & \textbf{TPR(\%)} & \textbf{FPR(\%)}  \\ 
\midrule
XGBoost + SMOTE~\cite{M-SMOTE}             & $77.1 \pm 0.01$    & $10.4 \pm 0.01$ \\ 
% \hline
XGBoost + AdaSyn-SMOTE~\cite{AdaSyn-SMOTE} & $77.0 \pm 0.01$    & $10.6 \pm 0.01$ \\ 
% \hline
XGBoost + cGAN~\cite{cGAN}                 & $78.5 \pm 0.01$    & $ 9.4 \pm 0.01$ \\
% \hline
XGBoost + DOS~\cite{DOS}                   & $78.6 \pm 0.01$    & $ 9.3 \pm 0.01$ \\
XGBoost + GAMO~\cite{GAMO}                 & $\mathbf{78.9 \pm 0.01}$ & $\mathbf{9.1 \pm 0.01}$ \\
% \hline
\bottomrule
\end{tabular}
\end{table}
\subsection{\textbf{Data Sparsity \& Imputation Method}}
Missing values are very common among most of the tabular datasets like ours. There are many methods available to impute missing values in dataset. However, if the degree of sparsity is high and the missing values are not imputed with high accuracy, it takes a toll on the generalization error of the model. An intuitive reason behind this is the fact that inaccurate imputation of data with high degree of sparsity, significantly alters the distribution of the data after imputation. It eventually results in the model learning a distribution which is significantly different from the ground-truth of the distribution. We observe that complex model-based imputation methods, such as MINWAE~\cite{MINWAE}, yield better true and false positive rate from the same model in comparison to simple mean or median imputation methods.

\section{Conclusion}\label{sec:conclusion}
In this paper, we introduce an ML based change risk assessment system which aims to bridge the gap between model-based assessment of change risks and the assessment of the domain experts. We also elaborate on how this system creates business impact. In near future, we will explore an active-learning based framework to leverage expert's feedback more optimally.

\bibliographystyle{unsrtnat}
\bibliography{references}  %%% Uncomment this line and comment out the ``thebibliography'' section below to use the external .bib file (using bibtex) .





% \end{thebibliography}

\end{document}